\documentclass[letterpaper, 10 pt, conference]{ieeeconf}  

\IEEEoverridecommandlockouts                              

\usepackage{graphics} 
\usepackage{epsfig} 
\usepackage{newtxmath} 
\usepackage{times} 
\usepackage{cite}
\usepackage{amsmath}

\usepackage{amssymb}
\usepackage{algorithmic}
\usepackage{subfigure}
\usepackage{textcomp}
\usepackage{xcolor}
\usepackage{booktabs}
\usepackage{multirow}
\usepackage{graphicx}
\usepackage[switch]{lineno}
\usepackage[linesnumbered,ruled,lined]{algorithm2e}
\usepackage[flushleft]{threeparttable}

\makeatletter
\newcommand\figcaption{\def\@captype{figure}\caption}
\newcommand\tabcaption{\def\@captype{table}\caption}
\makeatother

\title{\LARGE \bf
SOMTP: Self-Supervised Learning-Based Optimizer for MPC-Based Safe Trajectory Planning Problems in Robotics*
}

\author{Yifan Liu$^{1}$, You Wang$^{1, \dagger}$ and Guang Li$^{1}$
\thanks{*This work was supported by the Fundamental Research Funds for the Central Universities 226-2022-00086.}
\thanks{$^{1}$Yifan Liu, You Wang and Guang Li are with State Key Laboratory of Industrial Control Technology, Institute of Cyber Systems and Control, 
Zhejiang University, Hangzhou, 310027, China. 
{\tt\footnotesize yifanliu@zju.edu.cn, king\_wy@zju.edu.cn, guangli@zju.edu.cn}}
\thanks{$\dagger$ Corresponding author is You Wang.
}
}

\begin{document}

\maketitle
\thispagestyle{empty}
\pagestyle{empty}

\begin{abstract}
Model Predictive Control (MPC)-based trajectory planning has been widely used in robotics, and incorporating Control Barrier Function (CBF) constraints into MPC can greatly improve its obstacle avoidance efficiency. Unfortunately, traditional optimizers are resource-consuming and slow to solve such non-convex constrained optimization problems (COPs) while learning-based methods struggle to satisfy the non-convex constraints. In this paper, we propose SOMTP algorithm, a self-supervised learning-based optimizer for CBF-MPC trajectory planning. Specifically, first, SOMTP employs problem transcription to satisfy most of the constraints. Then the differentiable SLPG correction is proposed to move the solution closer to the safe set and is then converted as the guide policy in the following training process. After that, inspired by the Augmented Lagrangian Method (ALM), our training algorithm integrated with guide policy constraints is proposed to enable the optimizer network to converge to a feasible solution. Finally, experiments show that the proposed algorithm has better feasibility than other learning-based methods and can provide solutions much faster than traditional optimizers with similar optimality.
\end{abstract}

\section{Introduction}
Constrained optimization problems (COPs) have been widely used in many areas, such as trajectory planning in robotics, power systems, scheduling, and logistics. In some scenarios, COP may impose high demands on solving time. MPC is a special COP that generates a sequence of optimal control inputs and states. In recent years, MPC has gained extensive application in collision avoidance trajectory planning for autonomous driving and robots ~\cite{MPC_TP_Ji,MPC_TP_Ammour,CBF_MPC_Jian}. To achieve efficient collision avoidance trajectory planning for MPC, ~\cite{CBF_MPC} proposes an efficient CBF-MPC planner by incorporating discrete-time non-convex CBFs as constraints into MPC.  CBF-MPC is then widely used in the robot ~\cite{CBF_MPC_Jian,CBF_MPC_Wei,CBF_MPC_Yu}. CBF could enforce the system to avoid obstacles even when the reachable set is far away from the obstacles, as well as ensure forward invariance of the safe set ~\cite{CBF_MPC}. However, the addition of CBF also increases the complexity of the COP and thus lengthens the time required to solve it.

MPC-based trajectory planning problem needs to be solved within a single time step. And if it is resolved sooner, the robot can respond faster to external changes and uncertainty. Traditional optimization methods for such non-convex COPs usually require large computational resources and a long solving time. Learning-based methods can significantly reduce the solving time and have been applied to many optimization problems with or without simple constraints. However, in terms of MPC-based trajectory planning, fulfilling complex non-convex constraints becomes a major challenge for these methods ~\cite{CBF_MPC,ssl_dc3}. What's more, such sequential optimization problems may be more challenging since many robots' kinematic models, which act as equality constraints, are nonlinear. All in all, there is a demand for optimization methods to speed up optimization while strictly satisfying constraints.

In this paper, we propose SOMTP, a self-supervised learning-based optimizer, to solve the CBF-MPC-based safe trajectory planning problem with CBF constraints to avoid obstacles. SOMTP is highly concerned with constraint satisfaction and is capable of rapidly and effectively solving the CBF-MPC while ensuring feasibility. Specifically, first, we transcribe the original sequential optimization problem into nonlinear programming with only CBF constraints by employing the single-shooting method ~\cite{single_shooting}. Then, the following two algorithms are proposed to satisfy the CBF constraints: (1) the SLPG correction algorithm; and (2) the ALM-based training method with guide policy constraints. The SLPG correction procedure is inspired by DC3 in ~\cite{ssl_dc3} which can be incorporated into the training process. Unlike DC3, SLPG correction is to optimize a non-convex projection-type problem by performing sequential linearization, quadratic penalty method, and gradient descent. Compared with traditional projection-based method, the SLPG correction can deal with non-convex constraints and is more efficient in time and resource consuming. Instead of obtaining the true projection on the safe set, SLPG correction is intended to move the pre-solution from the optimizer network several steps closer to the safe set by generating a differentiable approximate solution within iteration limits. After that, we propose the ALM-based training method to achieve both optimality and feasibility by training the optimizer network and updating Lagrangian multipliers. Furthermore, in the ALM-based training method, the corrected solution from SLPG is again utilized as the guide policy to guide the network training. As a result, the optimizer network is expected to gradually converge to a solution of the CBF-MPC while constraints are satisfied. 

In summary, our key contributions are as follows:
\begin{itemize}
    \item SOMTP is a self-supervised learning-based optimization algorithm for the CBF-MPC-based trajectory planning problem and holds referential significance for other COPs and optimal control problems with non-convex constraints. Meanwhile, our strategy for CBF constraints is also meaningful for CBF-based safe RL.
    \item SLPG performs a differentiable correction, which can move the solution closer to the non-convex safe set and speed up the reduction of the violation. It is then treated as the guide policy in ALM-based training.
    \item ALM-based training algorithm is proposed to enable the optimizer network to converge to the feasible solution of the MPC. 
    \item Guide policy constraints are integrated into the ALM-based training algorithm to guide the learning process and accelerate convergence.
    \item Experiments demonstrate that our SOMTP has better feasibility than other learning-based approaches. Compared with traditional optimization methods, it can provide high-quality, feasible solutions much faster while still maintaining a similar level of optimality.
\end{itemize}

The paper is organized as follows: Section \uppercase\expandafter{\romannumeral2} presents the related works. Section \uppercase\expandafter{\romannumeral3} presents the background and preliminaries. In section \uppercase\expandafter{\romannumeral4}, SOMTP algorithm is proposed. Comparisons and experiments are given in \uppercase\expandafter{\romannumeral5}. Finally, conclusions are given in Section \uppercase\expandafter{\romannumeral6}.

\section{Related Works}
CBF-MPC trajectory planning belongs to the non-convex sequential COP and has been widely used in robotics. We intend to solve the above non-convex COP with learning-based methods. And our method's related works fall into the following three categories.

\textbf{Traditional optimization methods.} Before optimizing MPC, traditional methods need to transcribe such an optimal control problem into a nonlinear programming problem (which is also a COP) ~\cite{mpc_shooting}. The main optimization methods for such nonlinear COP are the interior point method (IPM) and sequential quadratic programming (SQP). ~\cite{numerical_optimization}. To obtain the COP's solution, these methods often require iterating multiple times and calculating the Hessian matrix, both of which can be time-consuming and resource-consuming. Moreover, the ALM algorithm mentioned above is also a traditional optimizer for solving COP, which can obtain solution and its associated Lagrangian multipliers through iterative methods while avoiding numerical instabilities ~\cite{ALM}.

\textbf{Learning-based methods.} Learning-based optimization methods are mainly divided into two types, i.e., supervised learning methods (SLMs) and self-supervised learning methods (SSLMs). SLMs concern training models that can map a problem instance's representation to the target solutions from a traditional optimizer ~\cite{CO_survey}. These methods require the use of a traditional optimizer to generate the target optimal solutions and pay much attention to the errors (e.g., MSE or MAE) between the network's outputs and the target solutions ~\cite{sl_Surrogate,SL_data,SL_Zamzam,sl_lag_dual}. SSLMs don't need additional traditional optimizers. Instead, they can directly train the optimizer network using the objective function and constraint violations ~\cite{ssl_pdl}. One of the SSLMs is to add constraint violations to the loss function for gradient descent. 
On top of this, the DC3 algorithm proposed in ~\cite{ssl_dc3} incorporates a correction process for inequality constraints during training, the details of which can be seen in \uppercase\expandafter{\romannumeral3}. And ~\cite{ssl_dc3_mpc} directly employs the DC3 to solve the MPC problem. DC3 is the most similar work to ours, with the following three major differences between our SOMTP and DC3: (1) We propose a differentiable SLPG correction for non-convex CBF constraints that can provide better direction for correction and push the pre-solution closer to the safe set; (2) The ALM-based training method can update associate Lagrangian multipliers and quadratic penalty terms during the training procedure; (3) To enhance the efficiency of ALM-based training, guide policy constraints are incorporated with SLGP correction as the guide policy. Besides, ~\cite{ssl_pdl} proposes PDL that uses primal-dual learning to alternately train two networks, one the optimizer network and one the Lagrangian multipliers' network.

\textbf{Reinforcement learning (RL).} Safe RL expects to handle constraints during training, which has given us some inspiration. Some directly enforce constraints by projecting pre-solutions onto a safe set using convex optimization layers (e.g., ~\cite{diff_qp}) in the case of general convex constraints ~\cite{rl_proj_cmdp,rl_cbf_proj_qp}. However, these projection-based approaches are inefficient since they often require a large amount of computing resources during training. What's more, the projection-type problem may be infeasible when the initial value is very distant from the safe set. Different from projection-based methods, our SLPG correction focuses on non-convex constraints and is intended to move the pre-solution several steps closer to the safe set within iteration limits. What's more, in recent years, CBF has gained popularity in MPC and safe RL methods due to the fact that it ensures the forward invariance of the safe set ~\cite{rl_cbf,rl_cbf_dmpc}.

\section{Background}
Trajectory planning aims to plan a trajectory for the robot from its initial position to its target state while avoiding obstacles and satisfying kinematic constraints. In recent years, optimization-based trajectory planning for mobile robotics like autonomous vehicles has been studied in many works. And the proposed algorithm in this study is intended to solve the aforementioned optimization-based trajectory planning problem with a learning-based method. The necessary preliminaries are as follows:

\subsection{CBF-MPC Trajectory Planning}
\begin{figure}[tb]
    \centering
    \includegraphics[width=0.7\linewidth]{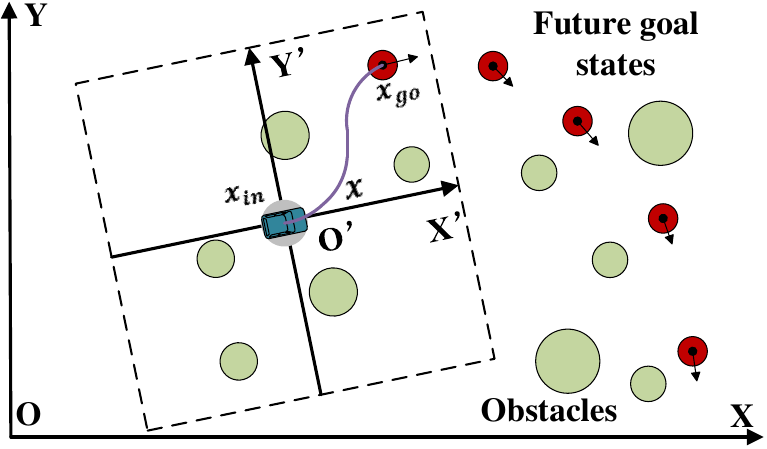}
    \caption{Frames of the robotic system.}
    \label{robot_frame}
    \vspace{-0.2cm}
\end{figure}

In this study, we consider a 2-D mobile robot with the following nonlinear kinematic model:
\begin{equation}
\boldsymbol{x_{k+1}} = \boldsymbol{f}(\boldsymbol{x_k}, \boldsymbol{u_k})
\label{eq1}
\end{equation}
where the states and controls of the robotic system at time-step k are $\boldsymbol{x_k} \in \mathbb{R}^{3}$ and $\boldsymbol{u_k} \in \mathbb{R}^{n_u}$, respectively. $\boldsymbol{x_k}=\begin{bmatrix} X & Y &\phi \end{bmatrix}$, where $X$ and $Y$ represent the coordinates in the local frame O'X'Y' in Fig.~\ref{robot_frame}. $\phi$ represents the yaw angle of the robot in the same frame.

Assume the robot's initial state and goal state are $\boldsymbol{x_{in}}$ and $\boldsymbol{x_{go}}$, respectively. And there are $n_{obs}\geq 0$ obstacles in the local cost map. To avoid the obstacles, we use $\mathcal{S}$ as a safe set, which is the superlevel set of a continuously differentiable function ${H}$ : $\mathcal{X} \subset \mathbb{R}^{3}$:
\begin{equation}
\mathcal{S} = \{\boldsymbol{x}\in\mathcal{X}, \boldsymbol{H}(\boldsymbol{x})>0\}
\label{eq2}
\end{equation}
where ${H}$ is the control barrier function (CBF) on $\mathcal{S}$ if there exists $\gamma\in(0, 1]$ such that for all $x\in\mathcal{S}$, ${H}$ satisfies
\begin{equation}
\sup_{\boldsymbol{u_k}} \;\; \Delta H(\boldsymbol{x_k}, \boldsymbol{u_k}) = H(\boldsymbol{x_{k+1}}) - H(\boldsymbol{x_k}) \geq -\gamma H(\boldsymbol{x_k})
\label{eq3}
\end{equation}

Given a CBF $H$, we can define the following set to guarantees the
forward invariance of $\mathcal{S}$ and the robots' safety.
\begin{equation}
G_{cbf} = \{\boldsymbol{u_k}\in\mathcal{U}, -\Delta H(\boldsymbol{x_k}, \boldsymbol{u_k}) - \gamma H(\boldsymbol{x_k}) \leq 0\}
\label{eq4}
\end{equation}

After combining the above discrete-time CBF constraints, the loss function of the CBF-MPC trajectory planner is obtained which can be seen as follows:
\begin{align}
\min_{\boldsymbol{u}(\cdot),\; \boldsymbol{x}(\cdot)} & && J(\cdot)=\sum_{k=0}^N
\left\|\boldsymbol{x_k}-\boldsymbol{{x}_{go}}\right\|^2_{\boldsymbol{Q}}+\sum_{k=0}^{N-1} \left\|\boldsymbol{u_k}\right\|^2_{\boldsymbol{R}} 
\label{eq5}\\
\text{s.t.} & &&
\boldsymbol{{x}_{k+1}}=\boldsymbol{f}\left(\boldsymbol{x_k},\;\; \boldsymbol{u_k}\right), \tag{5a} \label{eq5a}\\
& && \boldsymbol{x_0}=\boldsymbol{x_{in}},
\tag{5b} \label{eq5b}\\
& && \boldsymbol{u_k}\in\left[
\boldsymbol{u}_{min},\;\; \boldsymbol{u}_{max}\right], \tag{5c} \label{eq5d} \\
& &&-\Delta H(\boldsymbol{x_k}, \boldsymbol{u_k}|{\boldsymbol{O_j}}) - \gamma H(\boldsymbol{x_k}|{\boldsymbol{O_j}}) \leq 0 \tag{5d} \label{eq5e}
\end{align}
where $N$ is the prediction horizon, ${\boldsymbol{Q}}$ and ${\boldsymbol{R}}$ are definite weighting coefficient matrix. The number of \eqref{eq5e} is $N \cdot n_{obs}$. To help represent these obstacles, we use $\boldsymbol{O_j}=\begin{bmatrix} X_{o, j} & Y_{o, j} & R_{o, j}\end{bmatrix}, j\in [0, n_{obs})$, to represent the $j$-th obstacles, where $X_{o,j}$ and $Y_{o, j}$ represent the coordinates of the obstacle in the local frame O'X'Y' in Fig.~\ref{robot_frame}. $R_{o, j}$ represents the radius of the obstacle's smallest circumscribed circle. For the CBF function $H(\cdot)$ in \eqref{eq5e}, we employ the following function to simplify the obstacle avoidance problem:
\begin{equation}
    {H}\left(\boldsymbol{x}|\boldsymbol{O_j}\right)=(\boldsymbol{x}[0]-X_{o, j})^2 + (\boldsymbol{x}[1]-Y_{o, j})^2 - (R_{o, j}+R+l_{ex})^2
\label{eq6}
\end{equation}
where $R$ represents the radius of the robot and $l_{ex}$ is the expansion length of the obstacle. And the discrete-time CBF in \eqref{eq5e} is generally non-convex according to ~\cite{CBF_MPC}.

The above CBF-MPC belongs to the non-convex optimal control problem and needs to be solved using some nonlinear optimization algorithms such as SQP and IPM. However, these traditional optimization methods often require large computing resources and a long time to work.

\subsection{DC3}
DC3 is a learning-based method that intends to solve COPs by directly integrating two processes, i.e., equality completion and inequality correction, into the training procedure ~\cite{ssl_dc3}. Consider the following optimization problem as an instance:
\begin{equation}
\min_{\boldsymbol{y}} f_x(\boldsymbol{y}),\;\;\text{s.t.}\;\;h_x(\boldsymbol{y})=0, \;\;g_x(\boldsymbol{y})\leq0
\label{eq7}
\end{equation}
where $\boldsymbol{x}$ is the problem data and $\boldsymbol{y}\in\mathbb{R}^n$ is the corresponding solution. Loss function, equality, and inequality constraints are denoted by $f_x$, $h_x$ and $g_x$, respectively. Overall, DC3 will employ a neural network to output a partial set of solution $\boldsymbol{p}\in\mathbb{R}^m, m\leq n$. $\boldsymbol{p}$ will then be completed to a full set of solution $\boldsymbol{y_f}=\begin{bmatrix}
    \boldsymbol{p} & \phi_x(\boldsymbol{p})
\end{bmatrix}^T\in\mathbb{R}^n$, where they assume access to the function $\phi_x$ to satisfy the equality constraints. Next, the $\boldsymbol{y_f}$ will be corrected to $\boldsymbol{\hat{y}}$ by performing gradient descent on the violations of the inequality constraints during the correction process. Specifically, the correction is to modify the $\boldsymbol{y_f}$ with the equation $\boldsymbol{\hat{y}}=\boldsymbol{y_f}+\boldsymbol{\Delta y}=\begin{bmatrix}
    \boldsymbol{p}-\gamma \boldsymbol{\Delta p} & \phi_x(\boldsymbol{p})-\gamma\partial\phi_x(\boldsymbol{p})/\partial\boldsymbol{p}\boldsymbol{\Delta p}
\end{bmatrix}^T$
where $\boldsymbol{\Delta p} = \nabla_{\boldsymbol{p}}\|\text{ReLU}(g_x(\boldsymbol{y_f}))\|^2$ and $\gamma$ is the constant learning rate.

After the above procedures, the overall network will be trained using backpropagation on the following loss function:
\begin{equation}
l_{soft} = f_x(\boldsymbol{\hat{y}}) + \lambda_h \|h_x(\boldsymbol{\hat{y}})\|^2 + \lambda_g\|\text{ReLU}(g_x(\boldsymbol{\hat{y}}))\|^2
\label{eq8}
\end{equation}
where $\lambda_h,\lambda_g>0$ are hand-tuned hyper-parameters. 

\begin{figure}[tb]
    \centering
    \includegraphics[width=0.8\linewidth]{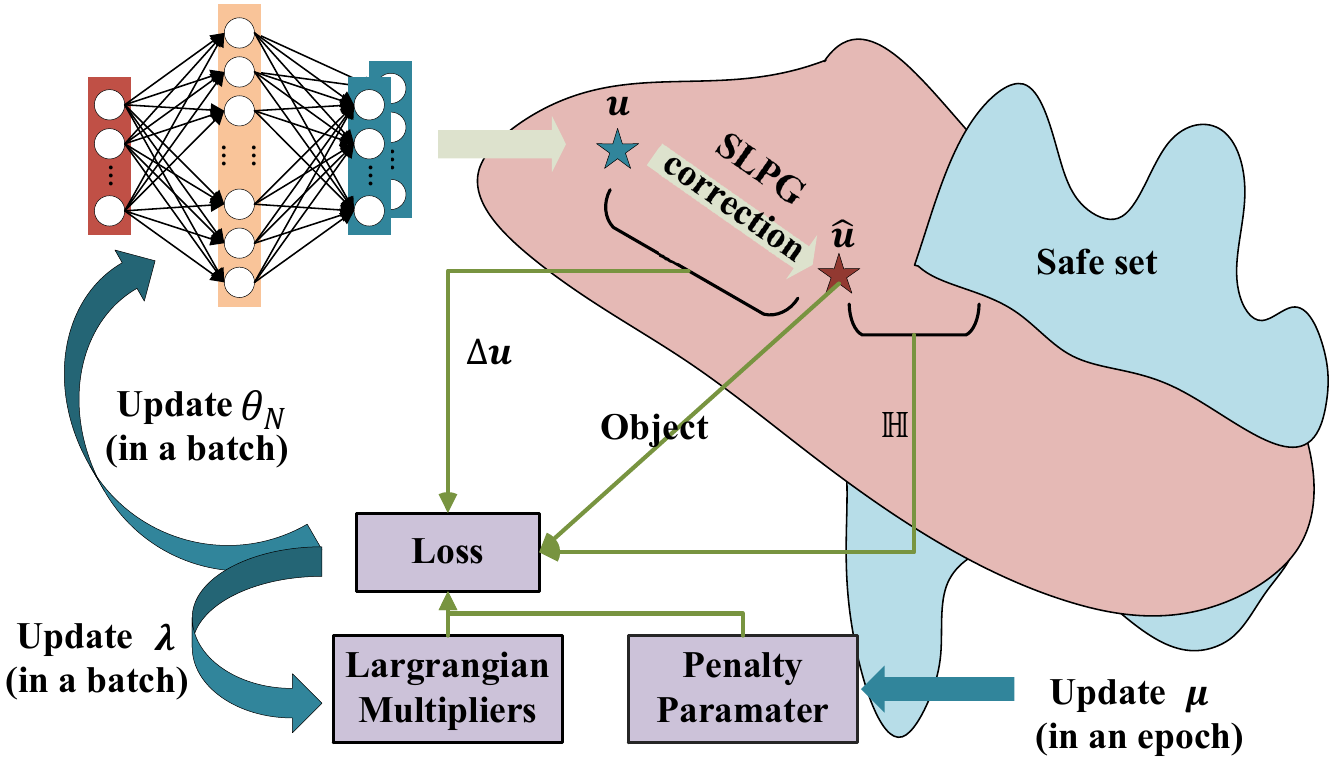}
    \caption{The structure of SOMTP algorithm.}
    \label{fig_somtp}
    \vspace{-0.2cm}
\end{figure}

\section{Algorithm}
To achieve safe trajectory planning, we employ the MPC planner with CBF to keep the robot away from obstacles. As it can be seen, the optimization problem in \eqref{eq5} is a complex, non-convex, nonlinear, sequential optimal control problem with nonlinear equality and inequality constraints. To solve the aforementioned problem, inspired by optimization algorithms, safe RL algorithms, and the learning-with-correction framework in DC3, we finally propose the learning-based optimization algorithm SOMTP. The structure of SOMTP can be seen in Fig.~\ref{fig_somtp}.

\subsection{Problem Transcription}
Before solving the problem in \eqref{eq5}, states and obstacles need to be transformed to the local coordinate system, with the current state as the zero point, and thus the initial state $\boldsymbol{x_{0}}=\boldsymbol{x_{in}}=\begin{bmatrix}0&0&0\end{bmatrix}$ can be seen as known. According to \eqref{eq5a} and \eqref{eq5b}, the sequential trajectory with respect to the control inputs $\boldsymbol{u}$ can be simplified as:
\begin{equation}
\boldsymbol{x_{k}} = \boldsymbol{F}(\boldsymbol{u_0}, \dots, \boldsymbol{u_{k-1}}), k\geq1
\label{eq9}
\end{equation}

After employing the above single-shooting method, the trajectory optimization problem in \eqref{eq5} can be transcribed as a non-linear programming problem in \eqref{eq10} ~\cite{single_shooting}.
\begin{align}
\min_{\boldsymbol{u}(\cdot)} & && \widehat J(\boldsymbol{u}, \boldsymbol{{x}_{go}})
\label{eq10}\\
\text{s.t.}
& && \boldsymbol{u_k}\in\left[
\boldsymbol{u}_{min},\;\; \boldsymbol{u}_{max}\right], k\in[0,N) \tag{10a} \label{eq10a} \\
& && H^{cbf}_{k,j}(\boldsymbol{u_0},\dots,\boldsymbol{u_k}, \boldsymbol{O_j})\leq 0, j\in[0,n_{obs})\tag{10b} \label{eq10b}
\end{align}
where $\widehat J(\cdot)=\sum_{k=1}^N
\left\|\boldsymbol{F}(\boldsymbol{u_{0:k-1}})-\boldsymbol{{x}_{go}}\right\|^2_{\boldsymbol{Q}}+\sum_{k=0}^{N-1} \left\|\boldsymbol{u_k}\right\|^2_{\boldsymbol{R}} $, $H^{cbf}_{k,j}(\cdot)=-\Delta H(\boldsymbol{F}(\boldsymbol{u_{0:k-1}}), \boldsymbol{u_k}|{\boldsymbol{O_j}}) - \gamma H(\boldsymbol{F}(\boldsymbol{u_{0:k-1}})|{\boldsymbol{O_j}})$.

\begin{figure}[tb]
    \centering
\includegraphics[width=0.8\linewidth]{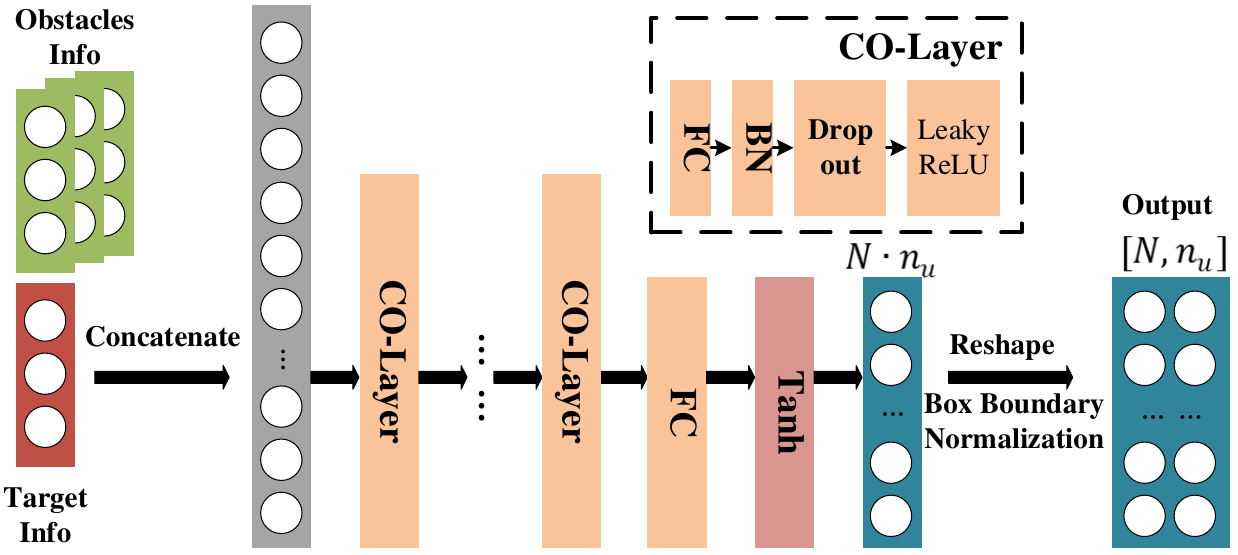}
    \caption{The structure of the network. Each CO-Layer in the network has the structure in the upper right corner of the figure.}
    \label{nn}
\end{figure}

Take a neural network denoted as $\theta_N$ to be the solver of the optimization problem \eqref{eq10}, the structure of which is shown in Fig.~\ref{nn}. The output of the final activation layer \text{tanh} will be bounded within the range of [-1,1]. This output will then be forwarded to the final layer, which applies box boundary normalization to the result to rescale it to the range in \eqref{eq10a}. And in this way, the problem in \eqref{eq10} will be transcribed to the following optimization problem \eqref{eq11}
\begin{align}
\min_{\boldsymbol{u}=\boldsymbol{\pi}_{\theta_N}} & && \widehat J(\boldsymbol{u}, \boldsymbol{{x}_{go}})
\label{eq11}\\
\text{s.t.}
& && H^{cbf}_{k,j}(\boldsymbol{u_0},\dots,\boldsymbol{u_k}, \boldsymbol{O_j})\leq 0, k\in[0,N), j\in[0,n_{obs})\tag{11a} \label{eq11a}
\end{align}
where $\boldsymbol{\pi}_{\theta_N}$ represents the result of the $\theta_N$. 

It is obvious that, following the transcription, the only constraints that require consideration in \eqref{eq11} are the CBF constraints denoted as \eqref{eq11a}. To deal with the CBF constraints, we implement the following two policies: the SLPG correction and the policy-guided ALM-based training method.

\subsection{SLPG Correction}
$\boldsymbol{\pi}_{\theta_N}$ cannot directly guarantee the CBF constraints. In DC3, they insert an inequality correction procedure that can hopefully pull the results closer to or within the safe set by performing gradient descent on the violations of the in-equality constraints. In terms of the CBF constraints in \eqref{eq11a}, the correction procedure in DC3 is to repeat the following:

\begin{equation}
\boldsymbol{\hat{u}_{dc3}} = \boldsymbol{u} - \gamma_{d}\nabla_{\boldsymbol{u}} \sum\|\text{ReLU}(H^{cbf}_{k,j})\|^2
\label{eq12}
\end{equation}
where $\boldsymbol{\hat{u}_{dc3}}$ is the corrected results with DC3-correction and $\gamma_{d}$ is the associated step-size, which is man-tuned ~\cite{ssl_dc3}.

However, the \eqref{eq11} is a complex non-convex problem with nonlinear 
implicit function \eqref{eq9}, and the CBF constraints show highly nonlinear and non-convex features. Therefore, the gradient descent-based correction method in DC3 may fail to find a suitable direction for correction. What's more, the corrected results $\boldsymbol{\hat{u}_{dc3}}$ may conflict with the constraints in \eqref{eq10a}.

Several safe RL methods employ projection algorithms to project the results to the safe set constraints during training procedures ~\cite{rl_proj_cmdp,rl_cbf_proj_qp}. Here, we also expect to design a projection-type correction procedure to deal with the CBF constraints. Define $\boldsymbol{\Delta u}$ as the input's correction value, and it can be obtained from the following optimization problem:
\begin{align}
\min_{\boldsymbol{\Delta u}(\cdot)} & && \sum_{k=0}^{N-1} \left\|\boldsymbol{\Delta u_k}\right\|^2_{\boldsymbol{R}} 
\label{eq13}\\
\text{s.t.}
& && 
\boldsymbol{x_{k}^c} = \boldsymbol{F}(\boldsymbol{u_{0:k-1}}+\boldsymbol{\Delta u_{0:k-1}})\tag{13a} \label{eq13a}\\
& && \boldsymbol{u_k}+\boldsymbol{\Delta u_k}\in\left[
\boldsymbol{u}_{min},\;\; \boldsymbol{u}_{max}\right], \tag{13b} \label{eq13b} \\
& &&H^{cbf}_{k,j}(\boldsymbol{u_{0:k}}+\boldsymbol{\Delta u_{0:k}}, \boldsymbol{O_j})\leq 0, j\in[0,n_{obs}) \tag{13c} \label{eq13c}
\end{align}

Moreover, the projection problem in the current safe RL algorithm is equivalent to a quadratic programming (QP) problem with linear or convex constraints. However, the problem \eqref{eq13} is not a QP problem at all with nonlinear and non-convex constraints, and solving this optimization problem will consume much time.

Inspired by SQP and SLQP algorithms that solve complex non-convex problems by solving a sequential of QP problem, we perform a first-order Taylor expansion on the constraints in \eqref{eq13} and optimize several QP sub-problems. The first-order Taylor expansion is performed at the points $(\boldsymbol{u}, \boldsymbol{x})$ which are the results from $\boldsymbol{\pi}_{\theta_N}$ . Thus, we can obtain a QP-type sub-problem as follows:
\begin{align}
\min_{\boldsymbol{\Delta u}(\cdot)} & && \sum_{k=0}^{N-1} \left\|\boldsymbol{\Delta u_k}\right\|^2_{\boldsymbol{R}} 
\label{eq14}\\
\text{s.t.}
& && 
\boldsymbol{x_{k}^c} = \boldsymbol{x_k}+[\nabla_{\boldsymbol{u_{0:k-1}}}\boldsymbol{F}]\boldsymbol{\Delta u_{0:k-1}}\tag{14a} \label{eq14a}\\
& && \boldsymbol{u_k}+\boldsymbol{\Delta u_k}\in\left[
\boldsymbol{u}_{min},\;\; \boldsymbol{u}_{max}\right], \tag{14b} \label{eq14b} \\
& &&H^{cbf}_{k,j}(\boldsymbol{u_{0:k}}, \boldsymbol{O_j})+[\nabla_{\boldsymbol{u_{0:k}}}{H^{cbf}_{k,j}}]\boldsymbol{\Delta u_{0:k}}\leq 0, j\in[0,n_{obs}) \tag{14c} \label{eq14c}
\end{align}
where \eqref{eq14a} can indeed be ignored since it has already been inserted into \eqref{eq14c} during the transcription.

However, the QP problem in \eqref{eq14} may be infeasible since the original solution from $\theta_N$ is insufficient during training, sometimes very distant from the safe set, and fails to meet the CBF constraints. This kind of situation happens frequently during the beginning stages of training. Since the purpose of the correction is to find a proper direction to reduce the vibrations of the CBF constraints, we adopt the Quadratic Penalty Method ~\cite{Penalty} to transfer the CBF constraints to the loss function. Instead of solving \eqref{eq14}, we turn to solving the QP problem in \eqref{eq15} to pull the solution closer to the safe set.
\begin{align}
\min_{\boldsymbol{\Delta u}(\cdot)} & && \sum_{k=0}^{N-1} \left\|\boldsymbol{\Delta u_k}\right\|^2_{\boldsymbol{R}} + \lambda_c\sum_{k=0}^{N-1}\sum_{j=0}^{n_{obs}-1}\|\text{ReLU}(\mathcal{H}^{cbf}_{k,j}(\cdot))\|^2
\label{eq15}\\
\text{s.t.}
& && \boldsymbol{u_k}+\boldsymbol{\Delta u_k}\in\left[
\boldsymbol{u}_{min},\;\; \boldsymbol{u}_{max}\right], \tag{15a} \label{eq15a} 
\end{align}
where $\mathcal{H}^{cbf}_{k,j}$ is equivalent to \eqref{eq14c}, $\lambda_c>0$ denotes the penalty parameter. Employing $J_{corr}$ to represent the loss function in \eqref{eq15}.

In terms of the box constraints QP problem in \eqref{eq15}, we perform gradient descent on the $J_{corr}$ to obtain the suitable direction. The clamp function is also executed to ensure compliance with the box constraints. Details are provided below:
\begin{equation}
\boldsymbol{\Delta \hat{u}^{i+1}} = \boldsymbol{\Delta {u}^{i}} - \gamma_{c}\nabla_{\boldsymbol{\Delta u^i}} J_{corr}
\label{eq16}
\end{equation}
where the appropriate step size $\gamma_{c}$ is determined by the linear search method that will be proposed later. 
\begin{equation}
\boldsymbol{\Delta {u}^{i+1}} = \text{Clamp}(\boldsymbol{u} + \boldsymbol{\Delta \hat{u}^{i+1}}, \boldsymbol{u}_{min}, \boldsymbol{u}_{max})-\boldsymbol{u}
\label{eq17}
\end{equation}

Furthermore, it was suggested by ~\cite{qp_ddp_box_const} that naive clamping for box constraints could damage the direction of descent and convergence. To address this, we modify the Armijo condition in ~\cite{Armijo} to propose the Armijo-box condition in \eqref{eq18} for the box constraints in the linear search method for $\gamma_c$, ensuring that $\boldsymbol{u} + \boldsymbol{\Delta \hat{u}^{i+1}}$ stays as much inside the box constraints as possible. 
\begin{align}
&(Armijo(\cdot) \;\;\textbf{and} \;\;\boldsymbol{u} + \boldsymbol{\Delta \hat{u}^{i+1}}(\gamma_c) \in \left[\boldsymbol{u}_{min},\boldsymbol{u}_{max}\right] ) \notag\\ &\textbf{or}\;\; \text{\emph{iteration counter}} \geq \text{\emph{maximum iteration}}
\label{eq18}
\end{align}

\IncMargin{1em}
\begin{algorithm} [tb]
\SetKwInOut{Input}{Input}\SetKwInOut{Output}{Output}
\SetKwInOut{Init}{Init}
\SetKwFunction{SeqLinear}{SequentialLinear}
\SetKwFunction{Armijo}{Armijo-BoxLinearSearch}
\SetKwFunction{Clamp}{Clamp}
\SetKwFunction{Initial}{Init}
\SetKwFunction{Max}{Max} \SetKwFunction{ReLU}{ReLU}
\caption{SLPG Correction}
\label{SLPG} 
\Input{Solution from $\boldsymbol{\pi}_{\theta_N}$: $\boldsymbol{u}$. Obstacles: $\boldsymbol{O}$. Max correction steps: $n_m$. Max QP steps: $i_m$. Box constraints: $[\boldsymbol{u}_{min}, \boldsymbol{u}_{max}]$. $\lambda_c$.}
\Output{Corrected solution: $\boldsymbol{\hat u}$. $\boldsymbol{\Delta u}$}
\Init{$\boldsymbol{\Delta u}$}
$\boldsymbol{u^0} \leftarrow \boldsymbol{u}$\;
\For{$n \in [0, n_m)$}{
    $\SeqLinear(\boldsymbol{u^n}), \Initial(\boldsymbol{\Delta u^0})$\;
    \For{$i \in [0, i_m)$}{
        $\boldsymbol{d^i} \leftarrow \nabla_{\boldsymbol{\Delta u^i}} J_{corr}$\;
        $\gamma_c \leftarrow \Armijo$ with condition \eqref{eq18}\;
        $\boldsymbol{\Delta \hat{u}^{i+1}} \leftarrow \boldsymbol{\Delta {u}^{i}} - \gamma_{c}\boldsymbol{d^i}$; in \eqref{eq16}\;
        $\boldsymbol{\Delta {u}^{i+1}} = \Clamp(\boldsymbol{u^n} + \boldsymbol{\Delta \hat{u}^{i+1}}, \boldsymbol{u}_{min}, \boldsymbol{u}_{max})-\boldsymbol{u}$}
    $\boldsymbol{u^{n+1}} \leftarrow \boldsymbol{u^n} + \boldsymbol{\Delta u^{i_m}} \;\;|\;\; \boldsymbol{\Delta u}, \boldsymbol{\hat u} \leftarrow \boldsymbol{\Delta u} + \boldsymbol{\Delta u^{i_m}},\boldsymbol{u^{n+1}}$\;
    Break if $\Max(\ReLU(H^{cbf}_{k,j}(\boldsymbol{\hat u_{0:k}}, \boldsymbol{O_j}))$ is tiny
 }
\end{algorithm}
\DecMargin{1em} 

Following the aforementioned procedures, we finally present the SLPG correction approach, which includes the key steps, i.e., the \textbf{S}equential \textbf{L}inearization, the Quadratic \textbf{P}enalty method and the \textbf{G}radient Decent with modified linear search condition. Pseudocode of SLPG can be seen in Algorithm~\ref{SLPG}.

\subsection{Augmented Lagrangian-based Training Algorithm with Guide Policy Constraints}
After the SLPG correction, we obtained the corrected input $\boldsymbol{\hat{u}}$. By incorporating SLPG into the training process, the overall objective of the training can be defined as addressing a following optimization problem \eqref{eq19} which is equivalent to \eqref{eq11}. \eqref{eq19b} is also employed as the \textbf{guide policy constraints} that can convert SLPG into a sub-guide policy during the training.
\begin{align}
\min_{\boldsymbol{\hat u}=\text{SLPG}(\boldsymbol{\pi}_{\theta_N})} & && \widehat J(\boldsymbol{\hat{u}}, \boldsymbol{{x}_{go}})
\label{eq19}\\
\text{s.t.}
& && H^{cbf}_{k,j}(\boldsymbol{\hat{u}_{0:K}}, \boldsymbol{O_j})\leq 0, k\in[0,N), j\in[0,n_{obs})\tag{19a} \label{eq19a}\\
& &&\boldsymbol{\Delta u} = \boldsymbol{\hat u} - \boldsymbol{\pi}_{\theta_N} = 0 \tag{19b} \label{eq19b}
\end{align}

Thus, we can build the following loss function to train the policy network, which can be seen as the ALM of the \eqref{eq19}, 
\begin{align}
\mathcal{L}(\boldsymbol{\hat{u}}, \boldsymbol{\lambda},\boldsymbol{\mu})
=&\widehat J(\boldsymbol{\hat{u}}, \boldsymbol{{x}_{go}}) + \sum_{i=0}^{N}\sum_{j=0}^{n_{obs}}\lambda_{i,j}^c\mathbb{H}_{i,j}+\frac{\mu^c}{2}\sum_{i=0}^{N}\sum_{j=0}^{n_{obs}}\mathbb{H}_{i,j}^2 \notag \\
+&\boldsymbol{\lambda^{du}}\cdot  \text{abs}(\boldsymbol{\Delta u})+ \frac{\mu^{du}}{2} \|\ \boldsymbol{\Delta u}\|^2
\label{eq20}
\end{align}
where $\mathbb{H}_{i,j}=\text{ReLU}(H^{cbf}_{i,j}(\boldsymbol{\hat{u}_{0:i}}, \boldsymbol{O_j}))$, $\boldsymbol{\lambda} = \begin{bmatrix}\boldsymbol{\lambda^c} & \boldsymbol{\lambda^{du}}
\end{bmatrix}$ is the Lagrange multiplier and belongs to a positive set. $\boldsymbol{\mu} = \begin{bmatrix}\mu^c & \mu^{du}\end{bmatrix}$ represents the corresponding penalty parameter, with all values being positive. And $ \text{abs}(\cdot)$ computes the absolute value of each element. It is important to note that the presence of the penalty on $\boldsymbol{\Delta u}$ is due to an implicit expectation in the original problem. This expectation is to guide the original policy network $\theta_N$ to approach the corrected policy from SLPG in order to enable $\theta_N$ to directly generate a feasible solution, thus making \eqref{eq19} be equivalent to \eqref{eq11}. What's more, such policy-guided training can optimize the training direction and speed up convergence.

A gradient search is then performed on $\mathcal{L}(\boldsymbol{\hat{u}}, \boldsymbol{\lambda},\boldsymbol{\mu})$ to obtain the optimal tuple, which is shown as follows:
\begin{equation}
\theta_N^{n+1} = \theta_N^{n+1} - \eta_\theta \nabla_{\theta_N^n}\mathcal{L}(\boldsymbol{\hat{u}^n}, \boldsymbol{\lambda}^n,\boldsymbol{\mu})
\label{eq21}
\end{equation}
\begin{equation}
\lambda^{c,n+1}_{i,j} = \lambda^{c,n}_{i,j} + \mu^{c,m}\mathbb{E}_{{batch}}[ \mathbb{H}_{i,j}] 
\label{eq22}
\end{equation}
\begin{equation}
\boldsymbol{\lambda^{du,n+1}} = \boldsymbol{\lambda^{du,n}} + \mu^{du, m}\mathbb{E}_{{batch}}[ \text{abs}(\boldsymbol{\Delta u})] 
\label{eq23}
\end{equation}
where ${{batch}}$ denotes the training batch size and $\eta_{\theta}$ represents the learning rate. $n$ and $m$ denote the $n$-th training step and $m$-th epoch, respectively. What's more, the penalty parameters will be updated at the end of each epoch when the following condition occurs:
\begin{align}
\text{if}\;\; \mathbb{E}_{{epoch}}&[ \|\boldsymbol{\mathbb{H}}\|^2] < \beta_{c}/\epsilon_c: \;\; \beta_{c} = \mathbb{E}_{{epoch}}[ \|\boldsymbol{\mathbb{H}}\|^2] \notag\\
&\mu^{c,m+1} = \text{min}(\epsilon_c \cdot \mu^{c,m}, \mu^{c}_{max})
\label{eq24}
\end{align}
\begin{align}
\text{if}\;\; \mathbb{E}_{{epoch}}&[ \|\boldsymbol{\Delta u}\|^2] < \beta_{du}/\epsilon_{du}:\;\; \beta_{du} = \mathbb{E}_{{epoch}}[ \|\boldsymbol{\Delta u}\|^2] \notag\\
&\mu^{du,m+1} = \text{min}(\epsilon_{du} \cdot \mu^{du,m}, \mu^{du}_{max})
\label{eq25}
\end{align}
where $\epsilon_c, \epsilon_{du}>1$ are constants and ${epoch}$ denotes the training epoch. As for now, the overall training process has been presented, which can also be seen in Algorithm~\ref{SOMTP}.

\IncMargin{1em}
\begin{algorithm} [tb]
\SetKwInOut{Input}{Input}\SetKwInOut{Output}{Output}
\SetKwInOut{Init}{Init}
\SetKwFunction{SLPG}{Alg.SLPG}
\caption{SOMTP (During Training)}
\label{SOMTP} 
\Input{Dataset: $\mathbb{D}$; Penalty: $\boldsymbol{\mu}$; Step size: $\eta_{\theta}$; Penalty updating values: $\epsilon_c, \epsilon_{du}$; Max penalty: $\mu^{c}_{max}$, $\mu^{du}_{max}$}
\Output{Policy Network: $\theta_N$}
\Init{$\theta_N$, $\boldsymbol{\lambda}$, $\boldsymbol{\mu}$, $\beta_c$, $\beta_{du}$}
\For{$m \leftarrow 0$ \text{to} $M$}{
     Shuffle dataset $\mathbb{D}$\;
\For{$n \leftarrow 0$ \text{to} $N$}{
    From $\mathbb{D}$ choose Data $d$;$\;\;$ $\boldsymbol{u} \leftarrow \theta_N(d)$\;
    $\boldsymbol{\hat{u}},\boldsymbol{\Delta u} \leftarrow \SLPG(\boldsymbol{u}, d)$\;
    Calculate $\mathcal{L}(\boldsymbol{\hat{u}}, \boldsymbol{\lambda}, \boldsymbol{\mu})$ using \eqref{eq20}\;
    Update $\theta_N$ using \eqref{eq21}\;
    Update $\boldsymbol{\lambda}$ using \eqref{eq22} and \eqref{eq23}\; 
 }
    If the condition occurs, update $\boldsymbol{\mu}$ and $[\beta_c,\beta_{du}]$ using \eqref{eq24} and \eqref{eq25} 
 }
 Return $\theta_N$
\end{algorithm}
\DecMargin{1em}

\section{Experimental Results}

In this section, experiments are carried out to compare the effectiveness of SOMTP to baselines on CBF-MPC-based safe trajectory planning problem. The chosen experimental agent is the autonomous vehicle with the following kinematic model:
\begin{equation}
\boldsymbol{x_{k+1}} = 
\boldsymbol{f}(\boldsymbol{x_k}, \boldsymbol{u_k})=
\begin{bmatrix}
    X + v\cos\phi\text{d}t\\
    Y + v\sin\phi\text{d}t\\
    \phi + v\tan q/L\text{d}t
\end{bmatrix}
\notag
\end{equation}
where $\boldsymbol{u_k} = \begin{bmatrix}v & q\end{bmatrix}^T$. Specifically, we compare SOMTP against the following baselines:
\begin{itemize}
    \item \textbf{IPOPT}, \textbf{SQPmethod} (Traditional Optimizer): We employ the IPM-based optimizer IPOPT ~\cite{ipopt} and the SQP-based optimizer SQPmethod (with OSQP ~\cite{osqp} to solve the sub-QP problem) in CasADi \cite{casadi}. In addition, CasADi is also utilized as an algorithmic differentiation tool in both optimizers;
    \item \textbf{MSE}, \textbf{MAE}: trained to minimize the L2 norm error or the L1 norm error between the optimizer network’s outputs and the optimal solutions from the traditional optimizer (IPOPT);
    \item \textbf{Penalty}: trained to minimize a soft loss in \eqref{eq8};
    \item \textbf{DC3}: proposed in ~\cite{ssl_dc3};
    \item \textbf{PDL}: proposed in ~\cite{ssl_pdl};
    \item \textbf{ALM} (ablation study): SOMTP without SLPG correction at both train and test time;
    \item \textbf{SOMTP-w/o$\boldsymbol{\Delta u}$} (ablation study): SOMTP without guide policy constraints in \eqref{eq19b} so that the ALM loss in \eqref{eq20} will not contain the penalty on $\boldsymbol{\Delta u}$;
\end{itemize}

The same neural network is employed across all experiments, with the structure in Fig.~\ref{nn} and five CO-Layers. Each CO-Layer has a hidden layer fully connected with 2000 nodes, which is then followed by a dropout layer (rate 0.3). The optimizer network is trained using PyTorch ~\cite{pytorch}, and the training process is executed on a system equipped with a GeForce RTX 3080 GPU and an Intel Xeon 2.9GHz CPU. To train the neural network, we generate the dataset with a total of 1 million examples (with train/test/validation ratio 18/1/1). For each example, the goal state $\boldsymbol{x_{go}}$ and obstacles $\boldsymbol{O_j}$ are randomly generated within the local cost map (6 m $\times$ 6 m). Each example has three obstacles ($n_{obs}=3$), each with a circular shape and a radius $R_{o} \in [0,\;0.5]$ m. The prediction horizon $N$ is 20. ${\boldsymbol{Q}}$ and ${\boldsymbol{R}}$ are $diag([2.0, 2.0, 1.0])$ and $diag([1.0, 1.5])$, respectively. In addition, ($\gamma$, $R$,  $l_{ex}$) = (0.5, 0.3, 0.1), which is used in \eqref{eq5} and \eqref{eq6}. 

\subsection{Results on the Test Dataset}
\begin{figure*}
    \begin{minipage}[h]{0.75\textwidth}
        \centering
        \begin{threeparttable}[b]
        \makeatletter\def\@captype{table}\makeatother\caption{Results on test dataset (with 50000 instances).\tnote{1}}
        \label{result_test}
        \begin{tabular}{lrrrrr}
        \toprule        
        \textbf{Algorithm}& \textbf{Obj.} & \textbf{Mean CBF} & \textbf{Max CBF} & \textbf{Infe. (\%)} & \textbf{Time (ms)} \\
        \midrule
        IPOPT  & 165.31   & 0.0000        & 0.0000       & 0.00             & \textcolor{red}{83.676}    \\
        SQPmethod    & 170.72   & 0.0000    & 0.0010  & 0.35          & 115.18    \\
        \midrule
        \textbf{SOMTP} *             & 167.93   & \textbf{0.0000}        & \textbf{0.0252}  & \textbf{0.07}          & \textcolor{red}{\textbf{1.057}}    \\
        SOMTP,$\nleqslant$        & 167.92   & 0.0000        & 0.0555  & 0.11          & 0.727   \\
        SOMTP-w/o$\boldsymbol{\Delta u}$ & 165.57   & 0.0002   & 0.1537  & 0.64          & 1.841    \\
        SOMTP-w/o$\boldsymbol{\Delta u}$,$\nleqslant$ & 165.55   & 0.0005   & 0.1537  & 0.85          & 0.717   \\
        ALM               & 164.70   & 0.0003   & 0.1746  & 0.87          & \textbf{0.724}   \\
        DC3               & 163.23   & 0.0006   & 0.2091  & 1.54          & 1.037   \\
        DC3,$\nleqslant$  & 163.23   & 0.0007   & 0.2129  & 1.54          & 0.741   \\
        Penalty           & 162.88   & 0.0008   & 0.1792  & 1.74          & 0.728   \\
        PDL               & \textbf{160.53}   & 0.0027   & 0.1807  & 3.71          & 0.744   \\
        MSE               & 169.74   & 0.1007   & 0.2703  & 16.56         & 0.760   \\
        MAE               & 164.89   & 0.1015   & 0.2794  & 17.48         & 0.739   \\ 
        \bottomrule
    \end{tabular}
    \begin{tablenotes}
    \small
     \item $\nleqslant$ denotes the absence of correction phase (like SLPG correction) at test time.
     \item[1]  \;To address the speed discrepancy in algorithmic differentiation like the calculation of gradients, traditional optimizers and learning-based algorithms that necessitate correction procedures utilize CasADi to provide gradients during testing. 
     \item * SOMTP can provide the output 79× faster than IPOPT with similar optimality and a very low infeasibility rate. Furthermore, it is far more feasible than other learning-based algorithms and shows a lower infeasibility rate compared to the traditional optimizer SQPmethod.
   \end{tablenotes}
    \end{threeparttable}
    \end{minipage}
    \begin{minipage}[h]{0.23\textwidth}
        \centering
\includegraphics[width=3.5cm]{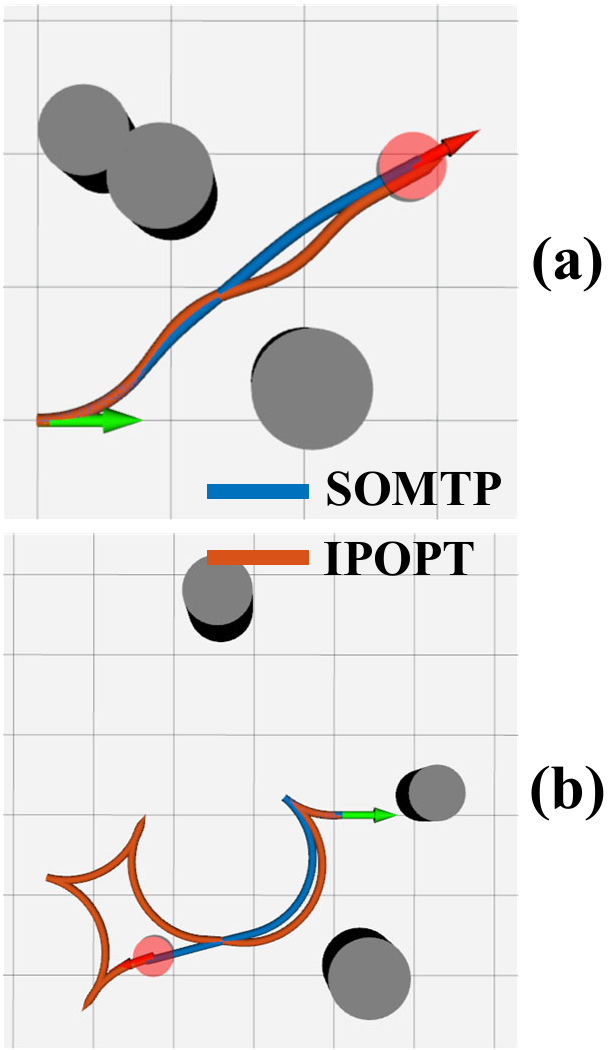} 
\makeatletter\def\@captype{figure}\makeatother\caption{Continues trajectory planning tasks on robot. Each grid is 1 m in width. The target area is denoted by a red circle. Green arrays represent the initial states, while red arrays represent the target states.}
        \label{obstacles_3}   
    \end{minipage}
\end{figure*}

Table.~\ref{result_test} compare the performance of the SOMTP algorithm with traditional optimizers and other learning-based methods on the test dataset. For our SLPG correction procedure, we use $(n_m^{train}, i_{m}^{train})$ = $(2,2)$ and $(n_m^{test}, i_{m}^{test})$ = $(10,2)$. The following five indicators are selected to compare the results:
\begin{itemize}
    \item Obj.: mean object value from the loss function in \eqref{eq5};
    \item Mean CBF: mean violations of the cumulative CBF non-convex constraints $\mathbb{E}_{test}[\sum_{i=0}^{N}\sum_{j=0}^{n_{obs}}\mathbb{H}_{i,j}]$;
    \item Max CBF: maximum violations on CBF constraints;
    \item Infe. (\%): infeasibility rate over the test dataset = (number of infeasible instances) / (total number of instances)$\times100\%$;
    \item Time (ms): mean time cost to solve.
\end{itemize}

\begin{figure*}[tb]
\begin{minipage}[h]{0.25\textwidth}
\makeatletter\def\@captype{table}\makeatother\caption{Result on robot in ten different planning tasks. }
    \begin{tabular}{lrr}
    \toprule        
    \textbf{Alg.}& \textbf{Suc.} & \textbf{Dist.} \\
    \midrule
    IPOPT & 90\% & 0.1112    \\
    SOMTP & 90\% & 0.2008 \\
    PDL   & 80\% & 0.2253 \\
    DC3   & 80\% & 0.2714  \\
    MAE   & 40\% & 0.0734   \\
    MSE   & 40\% & 0.0734  \\
    \bottomrule
    \end{tabular}

    \label{result_robot_table}
\end{minipage}
\begin{minipage}[h]{0.7\textwidth}
        \centering
   \includegraphics[height=3.4cm]{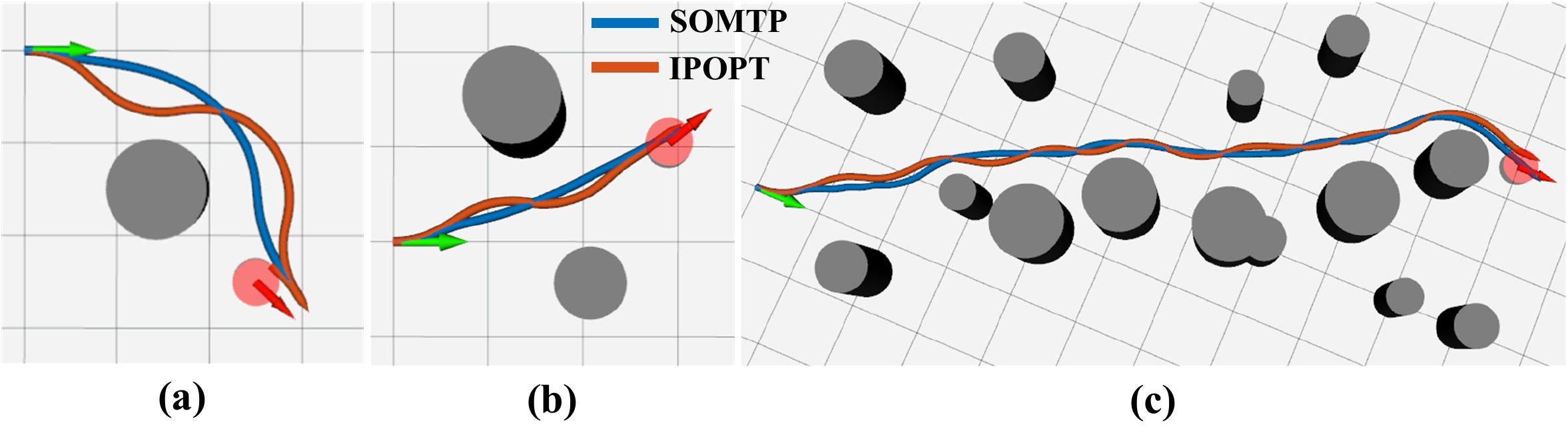}
 \caption{Apply optimizers to robot‘s trajectory planning tasks with variable quantities of obstacles. Each grid is 1 m in width. The target area is denoted by a red circle. Green arrays represent the initial states, while red arrays represent the target states.}
 \label{obstacles_multiple}
\end{minipage}
\end{figure*}

According to Table.~\ref{result_test}, SOMTP has a remarkably low infeasibility rate that is only surpassed by IPOPT. In addition, SOMTP is capable of providing the result 79× faster than IPOPT with similar optimality. The results also indicate that SSLMs are more feasible than SLMs. According to the results, the SLPG correction employed at test time can effectively reduce infeasibility. Nevertheless, the results of ALM and SOMTP-w/o$\boldsymbol{\Delta u}$$\nleqslant$ indicate that only applying SLPG correction to the training procedure without a guide policy constraint does not significantly contribute to its feasibility. By incorporating the guide policy constraints with $\boldsymbol{\Delta u}$ into the training procedure, the feasibility of SOMTP (or SOMTP$\nleqslant$) is significantly enhanced. This suggests that the guide policy constraints in \eqref{eq19b} and \eqref{eq20} can guide the learning process to a better point and accelerate convergence. What's more, though some baselines have very low Obj.,  they violate the obstacle constraints, and their strategies are likely to find the shortest trajectory between the current position and the target point in the obstacle-free task.

Overall, SOMTP achieves state-of-the-art (SOTA) in learning-based optimization algorithms for the CBF-MPC-based trajectory planning problem, with a much lower infeasibility rate, fewer violations of CBF constraints, and similar optimality. 

\subsection{Results on the Robot's Navigation}
We also apply our algorithm to the robot in 10 different tasks to compare the performance in continuous trajectory planning. The robot will repeatedly solve the CBF-MPC based trajectory planning problem in \eqref{eq5} in each time-step until it reaches the target area or encounters an obstacle. Results can be seen in Table.~\ref{result_robot_table} and Fig.~\ref{obstacles_3}. The task is considered successfully achieved when the robot reaches the target area. The success rate (the number of successful instances / the total number of instances) is denoted as [Suc.], whereas the average final weighted distance to the target in the success instances is represented as [Dist.]. Experiments show that during robot navigation, continuous trajectory planning and control will amplify the impact of infeasibility, causing frequent collisions with MAE and MSE. Over all the tasks, SOMTP achieves a high success rate and can sometimes complete tasks that IPOPT fails. Nevertheless, the SSLM-based optimizers ultimately achieve larger weighted distances than IPOPT. This could be attributed to the fact that SSLM-based methods have a higher tendency to train for tasks where the target is more away from the current position, which leads to larger losses.

Ultimately, to confirm the practicality and robustness of the SOMTP-based optimizer, we tested its performance on tasks with variable quantities of obstacles and long-distance planning. The figures in Fig.~\ref{obstacles_multiple} illustrate the practicality and robustness of the algorithm.

\section{Conclusions}
We propose SOMTP, a self-supervised learning-based optimization algorithm for CBF-MPC-based trajectory planning problem, which belongs to a complex non-convex COP. The core components of SOMTP consist of three parts: (1) the problem transcription can transcribe the problem into a neural network and satisfy most of the constraints; (2) the SLPG correction can pull the initial solution closer to the non-convex safe set and provide a guide policy for the following training process; and (3) the ALM-based training process with guide policy constraints integrated ensures that the network reaches a feasible point with regard to both feasibility and optimality. Experiments demonstrate that our SOMTP has significantly greater feasibility compared to previous learning-based algorithms, as well as being considerably faster than traditional optimizers while maintaining a similar level of optimality.

What's more, we believe that the SOMTP algorithm holds referential significance for other COPs and optimal control problems with non-convex constraints. And our strategy for CBF constraints is also meaningful for CBF-based safe RL. Therefore, in future work, we will try to extend SOMTP to other COPs as well as explore safe RL strategies based on the SOMTP algorithm.

\bibliographystyle{IEEEtran}
\bibliography{refer}

\end{document}